\documentclass{article}
\usepackage{spconf,amsmath,graphicx}

\usepackage{mwe} % to get dummy images
\usepackage{booktabs}
\usepackage{multirow}
\usepackage{hyperref}
\usepackage{xcolor}
\usepackage{caption}

\title{MedMNIST Classification Decathlon: \\A Lightweight AutoML Benchmark for Medical Image Analysis}

\name{Jiancheng Yang\textsuperscript{1,2} \quad Rui Shi\textsuperscript{1} \quad Bingbing Ni\textsuperscript{1,2}}
\address{\textsuperscript{1}Shanghai Jiao Tong University, Shanghai 200240, China
\\
\textsuperscript{2}MoE Key Lab of Artificial Intelligence, AI Institute, Shanghai Jiao Tong University
\\
{\tt\small \{jekyll4168,shi-rui,nibingbing\}@sjtu.edu.cn}}

\begin{document}

\twocolumn[{
\renewcommand\twocolumn[1][]{#1}
\maketitle

\begin{center}
    \vspace{-20px}
    \centering
    \includegraphics[width=0.85\textwidth]{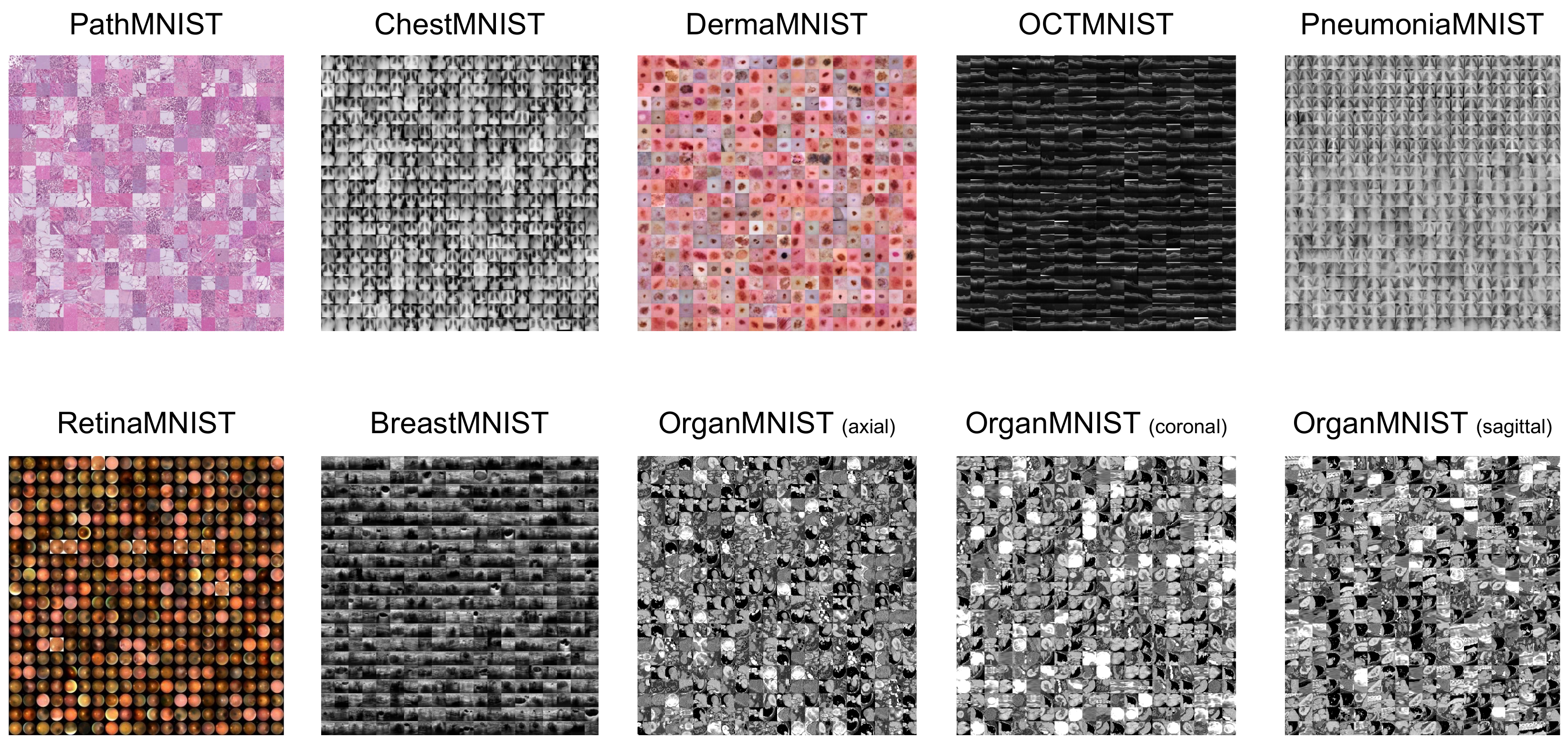}
    \captionof{figure}{\textbf{MedMNIST Classification Decathlon.} MedMNIST is a collection of 10 pre-processed medical image datasets with Creative Commons (CC) Licenses. It is designed to be educational, standardized, diverse and lightweight, which could be used as a rapid-prototyping playground and a multi-modal machine learning / AutoML benchmark in medical image analysis.}
    \label{fig:medmnist_overview}
\end{center}}]

\begin{abstract} 
We present \emph{MedMNIST}, a collection of 10 pre-processed medical open datasets. MedMNIST is standardized to perform classification tasks on lightweight $28\times 28$ images, which requires no background knowledge. Covering the primary data modalities in medical image analysis, it is diverse on data scale (from 100 to 100,000) and tasks (binary/multi-class, ordinal regression and multi-label). MedMNIST could be used for educational purpose, rapid prototyping, multi-modal machine learning or AutoML in medical image analysis. Moreover, \emph{MedMNIST Classification Decathlon} is designed to benchmark AutoML algorithms on all 10 datasets; We have compared several baseline methods, including open-source or commercial AutoML tools. The datasets, evaluation code and baseline methods for MedMNIST are publicly available at \url{https://medmnist.github.io/}. \footnote{Corresponding author: Bingbing Ni.}
\end{abstract}
\begin{keywords}
MedMNIST, AutoML, Classification, Multi-Modal, Benchmark, Decathlon.
\end{keywords}

\section{Introduction}
\label{sec:intro}

Medical image analysis, as an interdisciplinary area, is difficult to hand on for novices even from computer vision or clinical community, as it generally requires background knowledge from both sides. Especially analyzing multiple datasets with different modalities, it could be unfriendly as the datasets are generally nonstandard. On the other hand, although deep learning has been dominating the research and application in medical image analysis~\cite{litjens2017survey,shen2017deep}, it consumes large amounts of engineering labors to tune the deep learning models. As a result, automatic machine learning (AutoML)~\cite{faes2019automated} becomes increasingly important. However, there exists few benchmark for comparing AutoML in medical image classification.

To this end, we develop \emph{MedMNIST}, a collection of 10 pre-processed open medical image dataset. Following the MNIST~\cite{lecun2010mnist} datasets, the MedMNIST performs classification tasks on lightweight $28\times 28$ images, whose tasks cover primary medical image modalities and diverse data scales. By design, the MedMNIST is 
\begin{itemize}
    \item \textbf{Educational}: Our multi-modal data, from multiple open medical image datasets with Creative Commons (CC) Licenses, is easy to use for educational purpose.
    \item \textbf{Standardized}: Data is pre-processed into same format, which requires no background knowledge for users.
    \item \textbf{Diverse}: The multi-modal datasets covers diverse data scales (from 100 to 100,000) and tasks (binary/multi-class, ordinal regression and multi-label).
    \item \textbf{Lightweight}: The small size of $28\times 28$ is friendly for rapid prototyping and experimenting multi-modal machine learning and AutoML algorithms.
\end{itemize}

Inspired from Medical Segmentation Decathlon~\cite{simpson2019large}, we design \emph{MedMNIST Classification Decathlon}, as illustrated in Fig. \ref{fig:medmnist_overview}, to serve as a benchmark for AutoML in medical image classification. It evaluates the performance of AutoML algorithms on all 10 datasets, without any manual tuning. We have compared the performance of several baseline methods, including early-stopping ResNets~\cite{he2016deep}, open-source AutoML tools (auto-sklearn~\cite{NIPS2015_5872} and AutoKeras~\cite{jin2019auto}) and a commercial AutoML tool (Google AutoML Vision). We hope the benchmarking by MedMNIST Classification Decathlon could facilitate the future AutoML research in medical image analysis.

\section{MedMNIST Dataset}
\label{sec:dataset}

\begin{table*}
\small
\centering
\caption{{\bf An Overview of MedMNIST Dataset}.}
\label{tab:overview}
\begin{tabular}{@{}ccccccc@{}}
\toprule
Name & Source & Data Modality & Tasks (\# Classes / Labels) & \# Training & \# Validation & \# Test \\ \midrule
PathMNIST & \cite{10.1371/journal.pmed.1002730} & Pathology & Multi-Class (9) & 89,996 & 10,004 & 7,180 \\
ChestMNIST & \cite{wang2017chestxray} & Chest X-ray & Multi-Label(14) Binary-Class (2) & 78,468 & 11,219 & 22,433 \\
DermaMNIST & \cite{DBLP:journals/corr/abs-1803-10417,codella2019skin} & Dermatoscope & Multi-Class (7) & 7,007 & 1,003 & 2,005 \\
OCTMNIST & \cite{KERMANY20181122} & OCT & Multi-Class (4) & 97,477 & 10,832 & 1,000 \\
PneumoniaMNIST & \cite{KERMANY20181122} & Chest X-ray & Binary-Class (2) & 4,708 & 524 & 624 \\
RetinaMNIST & \cite{deepdr} & Fundus Camera & Ordinal Regression (5) & 1,080 & 120 & 400 \\
BreastMNIST & \cite{ALDHABYANI2020104863} & Breast Ultrasound & Binary-Class (2) & 546 & 78 & 156 \\
OrganMNIST\_Axial & \cite{DBLP:journals/corr/abs-1901-04056,8625393} & Abdominal CT & Multi-Class (11) & 34,581 & 6,491 & 17,778 \\
OragnMNIST\_Coronal & \cite{DBLP:journals/corr/abs-1901-04056,8625393} & Abdominal CT & Multi-Class (11) & 13,000 & 2,392 & 8,268 \\
OrganMNIST\_Sagittal & \cite{DBLP:journals/corr/abs-1901-04056,8625393} & Abdominal CT & Multi-Class (11) & 13,940 & 2,452 & 8,829 \\ \bottomrule
\end{tabular}
\end{table*}

The MedMNIST dataset consists of 10 pre-processed datasets from selected sources covering primary data modalities (e.g., X-ray, OCT, Ultrasound, CT), diverse classification tasks (binary/multi-class, ordinal regression and multi-label) and data scales (from 100 to 100,000). All source datasets are associated with Creative Commons (CC) Licenses or free licenses, which allows us to develop derivative datasets based on them, As depicted in Table \ref{tab:overview}, the diversity of dataset design could lead to that of task difficulties, which is desirable for an AutoML benchmark. We standardize each dataset by pre-processing and splitting it into training-validation-test subsets. The downsizing operation is implemented with Resize function provided by torchvision.transforms. We use the official data split from the source datasets if provided; or we split the whole / training datasets by 7:1:2 (train:val:test) / 9:1 (train:val) at the patient level.

\textbf{PathMNIST} is based on a prior study~\cite{10.1371/journal.pmed.1002730} for predicting survival from colorectal cancer histology slides, which provides a dataset (NCT-CRC-HE-100K) of $ 100,000 $ non-overlapping image patches from hematoxylin \& eosin stained histological images, and a test dataset (CRC-VAL-HE-7K) of $ 7,180 $ image patches from a different clinical center. 9 types of tissues are involved, resulting a multi-class classification task. We resize the source images of $3 \times 224 \times 224$ into $3 \times 28 \times 28$, and split NCT-CRC-HE-100K into training and valiation set with a ratio of 9:1.

\textbf{ChestMNIST} is based on NIH-ChestXray14 dataset~\cite{wang2017chestxray}, a dataset comprising $ 112,120 $ frontal-view X-ray images of $ 30,805 $ unique patients with the text-mined 14 disease image labels, which could be formulized as multi-label binary classification task. We use the official data split, and resize the source images of $1 \times 1024 \times 1024$ into $1 \times 28 \times 28$.

\textbf{DermaMNIST} is based on HAM10000~\cite{DBLP:journals/corr/abs-1803-10417,codella2019skin}, a large collection of multi-source dermatoscopic images of common pigmented skin lesions. The dataset consists of $ 10,015 $ dermatoscopic images labeled as 7 different categories, as a multi-class classification task. 
We split the images into training, validation and test set with a ratio of $ 7:1:2 $. The source images of $ 3 \times 600 \times 450 $ are resized into $ 3 \times 28 \times 28 $.

\textbf{OCTMNIST} is based on a prior dataset~\cite{KERMANY20181122} of $109,309$ valid optical coherence tomography (OCT) images for retinal diseases. 4 types are involved, leading to a multi-class classification task. 
We split the source training set with a ratio of $ 9:1 $ into training and validation set, and use its source validation set as the test set. The source images are single-channel, and their sizes are $ (384-1,536) \times (277-512) $; We center-crop the images and resize them into $ 1 \times 28 \times 28 $.

\textbf{PneumoniaMNIST} is based on a prior dataset~\cite{KERMANY20181122} of $ 5,856 $ pediatric chest X-ray images. The task is binary-class classification of pneumonia and normal. We split the source training set with a ratio of $ 9:1 $ into training and validation set, and use its source validation set as the test set. The source images are single-channel, and their sizes are $ (384-2,916) \times (127-2,713) $; We center-crop the images and resize them into $ 1 \times 28 \times 28 $.

\textbf{RetinaMNIST} is based on DeepDRiD~\cite{deepdr}, a dataset of $1,600$ retina fundus images. The task is ordinal regression for 5-level grading of diabetic retinopathy severity. We split the source training set with a ratio of $ 9:1 $ into training and validation set, and use the source validation set as test set. The source images of $ 3 \times 1,736 \times 1,824 $ are center-cropped and resized into $ 3 \times 28 \times 28 $.

\textbf{BreastMNIST} is based on a dataset~\cite{ALDHABYANI2020104863} of 780 breast ultrasound images. It is categorized into 3 classes: normal, benign and malignant. As we use low-resolution images, we simplify the task into binary classification by combing normal and benign as positive, and classify them against malignant as negative. We split the source dataset with a ratio of $ 7:1:2 $ into training, validation and test set. The source images of $ 1 \times 500 \times 500 $ are resized into $ 1 \times 28 \times 28 $.

\textbf{OrganMNIST\_\{Axial,Coronal,Sagittal\}} is based on 3D computed tomography (CT) images from Liver Tumor Segmentation Benchmark (LiTS)~\cite{DBLP:journals/corr/abs-1901-04056}. We use bounding-box annotations of 11 body organs from another study~\cite{8625393} to obtain the organ labels. Hounsfield-Unit (HU) of the 3D images are transformed into grey scale with a abdominal window; we then crop 2D images from the center slices of the 3D bounding boxes in axial / coronal / sagittal views (planes). The only differences of OrganMNIST\_\{Axial,Coronal,Sagittal\} are the views. The images are resized into $ 1 \times 28 \times 28 $ to perform multi-class classification of 11 body organs. 115 and 16 CT scans from the source training set are used as training and validation set, respectively. The 70 CT scans from the source test set are treated as the test set.

\section{Benchmark}

\begin{table*}[!htb]
	\caption{{\bf Overall performance of MedMNIST} in metrics of AUC and ACC, using ResNet-18 / ResNet-50~\cite{he2016deep} with resolution $28$ and $224$, auto-sklearn~\cite{NIPS2015_5872} , AutoKeras~\cite{jin2019auto} and Google AutoML Vision.}
	\label{tab:Results}
	\vspace{-10px}
	\begin{center}
		
		\begin{tabular}{@{}ccccccccccc@{}}
			\toprule
			\multirow{2}{*}{Methods} &
			\multicolumn{2}{c}{PathMNIST} &
			\multicolumn{2}{c}{ChestMNIST} &
			\multicolumn{2}{c}{DermaMNIST} &
			\multicolumn{2}{c}{OCTMNIST} &
			\multicolumn{2}{c}{PneumoniaMNIST} \\
			& AUC & ACC & AUC & ACC & AUC & ACC & AUC & ACC & AUC & ACC \\ \midrule
			ResNet-18 (28)~\cite{he2016deep}     & 0.972 & 0.844 & 0.706 & 0.947 & 0.899 & 0.721 & 0.951 & \bf0.758 & 0.957 & 0.843  \\
			ResNet-18 (224)~\cite{he2016deep}        & 0.978 & 0.860 & 0.713 & \bf0.948 & 0.896 & 0.727 & 0.960 & 0.752 & 0.970 & 0.861  \\
			ResNet-50 (28)~\cite{he2016deep}         & 0.979 & \bf0.864 & 0.692 & 0.947 & 0.886 & 0.710 & 0.939 & 0.745 & 0.949 & 0.857  \\
			ResNet-50 (224)~\cite{he2016deep}        & 0.978 & 0.848 & 0.706 & 0.947 & 0.895 & 0.719 & 0.951 & 0.750 & 0.968 & 0.896  \\
			auto-sklearn~\cite{NIPS2015_5872}         & 0.500 & 0.186 & 0.647 & 0.642 & 0.906 & 0.734 & 0.883 & 0.595 & 0.947 & 0.865  \\
			AutoKeras~\cite{jin2019auto}           & 0.979 & \bf0.864 & 0.715 & 0.939 & 0.921 & 0.756 & 0.956 & 0.736 & 0.970 & 0.918 \\
			Google AutoML Vision  & \bf0.982 & 0.811 & \bf0.718 & 0.947 & \bf0.925 & \bf0.766 & \bf0.965 & 0.732 & \bf0.993 & \bf0.941 \\
			\midrule
			\multirow{2}{*}{Methods} &
			\multicolumn{2}{c}{RetinaMNIST} &
			\multicolumn{2}{c}{BreastMNIST} &
			\multicolumn{2}{c}{OrganMNIST\_A} &
			\multicolumn{2}{c}{OrganMNIST\_C} &
			\multicolumn{2}{c}{OrganMNIST\_S} \\
			& AUC & ACC & AUC & ACC & AUC & ACC & AUC & ACC & AUC & ACC \\ \midrule
			ResNet-18 (28)~\cite{he2016deep}         & 0.727 & 0.515 & 0.897 & 0.859 & 0.995 & 0.921 & 0.990 & 0.889 & 0.967 & 0.762 \\
			ResNet-18 (224)~\cite{he2016deep}        & 0.721 & 0.543 & 0.915 & \bf0.878 & \bf0.997 & \bf0.931 & 0.991 & 0.907 & \bf0.974 & 0.777 \\
			ResNet-50 (28)~\cite{he2016deep}         & 0.719 & 0.490 & 0.879 & 0.853 & 0.995 & 0.916 & 0.990 & 0.893 & 0.968 & 0.746 \\
			ResNet-50 (224)~\cite{he2016deep}        & 0.717 & \bf0.555 & 0.863 & 0.833 & \bf0.997 & \bf0.931 & \bf0.992 & 0.898 & 0.970 & 0.770 \\
			auto-sklearn~\cite{NIPS2015_5872}        & 0.694 & 0.525 & 0.848 & 0.808 & 0.797 & 0.563 & 0.898 & 0.676 & 0.855 & 0.601 \\
			AutoKeras~\cite{jin2019auto}           & 0.655 & 0.420 & 0.833 & 0.801 & 0.996 & 0.929 & \bf0.992 & \bf0.915 & 0.972 & \bf0.803 \\
			Google AutoML Vision  & \bf0.762 & 0.530 & \bf0.932 & 0.865 & 0.988 & 0.818 & 0.986 & 0.861 & 0.964 & 0.706 \\
			\bottomrule
		\end{tabular}
		\vspace{-15px}
	\end{center}
\end{table*}

\begin{figure*}
	\centering
	\includegraphics[width=0.9\linewidth]{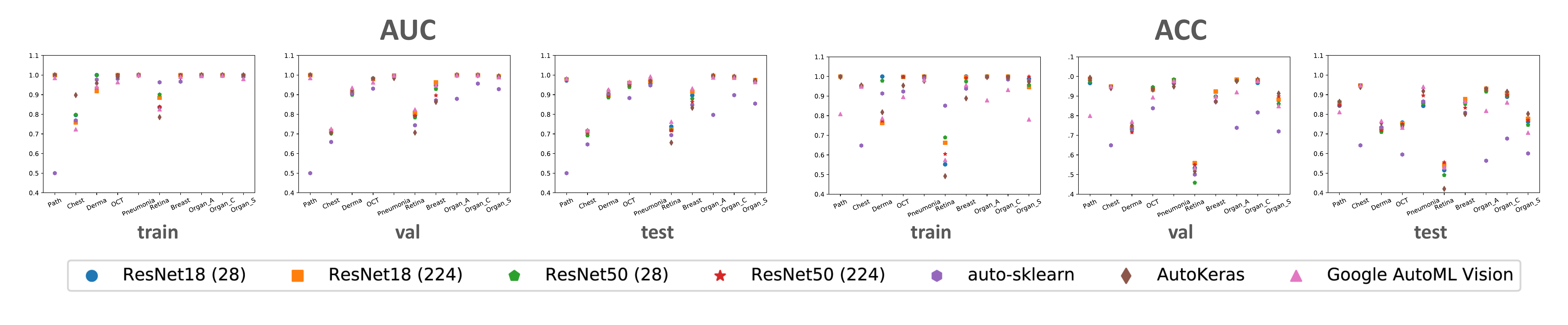}
	\caption{\textbf{Performance Analysis of Various Methods}, on the training, validation and test data splits in terms of AUC and ACC.}
	\label{fig:performance-analysis}
\end{figure*}

Inspired from Medical Segmentation Decathlon~\cite{simpson2019large}, we design MedMNIST Classification Decathlon, which aims at a lightweight AutoML benchmark for medical image analysis. It evaluates the algorithm performance on all 10 datasets without any manual tuning. Several methods are compared on the MedMNIST Classification Decathlon benchmark.

\subsection{Baseline Methods}

We first implement ResNets~\cite{he2016deep} with a simple early-stopping strategy on validation set as baseline methods. The input channel is $1$ for grey-scale dataset, and $3$ for triple-channel dataset. To farily compare with other methods, the input resolutions are $28$ or $224$ (resized from $28$) for the ResNet-18 and ResNet-50. The models are trained for 100 epochs, using a cross entropy loss and an SGD optimizer with a batch size of 128 and an initial learning rate $ 1 \times 10^{-3} $. 

\subsection{AutoML Methods}

We have also selected several AutoML methods on the MedMNIST Classification Decathlon: auto-sklearn~\cite{NIPS2015_5872} as the representative of open-source AutoML tools for typical statistical machine learning, AutoKeras~\cite{jin2019auto} as the representative of open-source AutoML tools for deep neural networks, and Google AutoML Vision as the representative of commercial black-box AutoML tools.

\textbf{auto-sklearn}~\cite{NIPS2015_5872} automatically searches the algorithms and hyper-parameters in scikit-learn \cite{pedregosa2011scikit} package.
We set time limits for search of appropriate models according to the dataset scale. the time limits are 2 hours for datasets with scale $<10,000$, 4 hours for those of $[10,000,50,000]$, and 6 hours for those $>50,000$.
We flatten the $28\times 28$ images into one dimension, and provide reshaped one-dimensional data of $784$ and the corresponding labels for auto-sklearn to fit.

\textbf{AutoKeras}~\cite{jin2019auto} based on Keras package~\cite{chollet2015keras} searches deep neural networks and hyper-parameters. For each dataset, we set $max\_trials$ to 20 and epochs to 20. It tries 20 different Keras models trained for 20 epochs each. We choose the best model based on the highest AUC score on validation set.

\textbf{Google AutoML Vision\footnote{\url{https://cloud.google.com/vision/automl/docs} (experimented on 2020-09-20).}} is a commercial AutoML tool provided on Google Cloud. We train Edge exportable models on Google AutoML Vision and export trained quantized models into TensorFlow Lite format to do offline inference. We set number of node hours of each dataset based on the data scale, 1 node hour for dataset with scale around $1,000$, 2 node hours for scale around $10,000$, and 3 node hours for scale around $100,000$. 

\subsection{Evaluation}
Area under ROC curve (AUC) and Accuracy (ACC) are used as the evaluation metrics. AUC is a threshold-free metric to evaluate the continuous prediction scores, while ACC evaluates the discrete prediction labels given threshold (or $\arg\max$). AUC is less sensitive to class imbalance than ACC. There is no severe class imbalance on our datasets, thus ACC could also serve as a good metric. Although there are many other metrics, we simply select AUC and ACC for the sake of simplicity and standardization of evaluation. We report the AUC and ACC of each dataset in this paper, while data users are encouraged to analyze the average performance over the 10 datasets to benchmark their AutoML methods.

\section{Results}

\subsection{Overall Performance}

The overall performance of the methods are reported in Table \ref{tab:Results}. Google AutoML Vision is well-performing in general, however it could not always win, even compared with the baseline ResNet-18 and ResNet-50. auto-sklearn performs poorly on most datasets, indicating that the typical statistical machine learning algorithms does not work well on our medical image datasets. AutoKeras performs well on datasets with large scales, however relatively worse on datasets with small scale.  Given the fact that no algorithm generalizes well on all 10 datasets, it could be interesting and practical to explore AutoML algorithms generalizing well on diverse data modalities, tasks and scales. 

\subsection{Performance Analysis on Data Splits}

We have also analyzed the overfitting / underfitting issues on the training, validation and test data splits. As illustrated in Fig. \ref{fig:performance-analysis}, algorithms tend to overfit the datasets with small scales. Google AutoML Vision controls the overfitting issues well, and severe overfitting is observed for auto-sklearn. We could infer that appropriate reductive bias for learning algorithms is crucial. It is also worthy exploring the regularization techniques on the MedMNIST, e.g., data augmentation, model ensemble, optimization algorithms.

\section{Conclusion}

We present MedMNIST, an \textbf{educational}, \textbf{standardized}, \textbf{diverse} and \textbf{lightweight} dataset for medical image classification, consisting of 10 pre-processed datasets. It is friendly for educational purpose and rapid prototyping multi-modal machine learning and AutoML algorithms. We also design MedMNIST Classification Decathlon, to serve as an AutoML benchmark for medical image classification. It evaluates the algorithm performance on all 10 datasets, without any manual tuning. We have compared the performance of several methods, including baseline early-stopping strategy, open-source AutoML tools and a commercial AutoML tool. The experiments indicate that no algorithms generalize well enough on all 10 datasets. We hope the benchmarking by MedMNIST Classification Decathlon could facilitate the future AutoML research in medical image analysis.

\section{Compliance with Ethical Standards}
\label{sec:ethics}
This research study was conducted retrospectively using human subject open-source data (details in Table \ref{tab:overview}). Ethical approval was not required as confirmed by the license attached with the open-access data.

\section{Acknowledgement}
This work was supported by National Science Foundation of China (U20B200011, 61976137). The authors would like to appreciate the SJTU Student Innovation Center for GPUs.

\bibliographystyle{IEEEbib}
\bibliography{refs}

\begin{thebibliography}{10}

\bibitem{litjens2017survey}
Geert Litjens, Thijs Kooi, et~al.,
\newblock ``A survey on deep learning in medical image analysis,''
\newblock {\em Medical image analysis}, vol. 42, pp. 60--88, 2017.

\bibitem{shen2017deep}
Dinggang Shen, Guorong Wu, and Heung-Il Suk,
\newblock ``Deep learning in medical image analysis,''
\newblock {\em Annual review of biomedical engineering}, vol. 19, pp. 221--248,
  2017.

\bibitem{faes2019automated}
Livia Faes, Siegfried~K Wagner, et~al.,
\newblock ``Automated deep learning design for medical image classification by
  health-care professionals with no coding experience: a feasibility study,''
\newblock {\em The Lancet Digital Health}, vol. 1, no. 5, pp. e232--e242, 2019.

\bibitem{lecun2010mnist}
Yann LeCun, Corinna Cortes, and CJ~Burges,
\newblock ``Mnist handwritten digit database,''
\newblock {\em ATT Labs [Online]. Available: http://yann.lecun.com/exdb/mnist},
  vol. 2, 2010.

\bibitem{simpson2019large}
Amber~L Simpson, Michela Antonelli, et~al.,
\newblock ``A large annotated medical image dataset for the development and
  evaluation of segmentation algorithms,''
\newblock {\em arXiv preprint arXiv:1902.09063}, 2019.

\bibitem{he2016deep}
Kaiming He, Xiangyu Zhang, et~al.,
\newblock ``Deep residual learning for image recognition,''
\newblock in {\em CVPR}, 2016, pp. 770--778.

\bibitem{NIPS2015_5872}
Matthias Feurer, Aaron Klein, et~al.,
\newblock ``Efficient and robust automated machine learning,''
\newblock in {\em Advances in Neural Information Processing Systems 28},
  C.~Cortes, N.~D. Lawrence, D.~D. Lee, M.~Sugiyama, and R.~Garnett, Eds., pp.
  2962--2970. Curran Associates, Inc., 2015.

\bibitem{jin2019auto}
Haifeng Jin, Qingquan Song, and Xia Hu,
\newblock ``Auto-keras: An efficient neural architecture search system,''
\newblock in {\em Proceedings of the 25th ACM SIGKDD International Conference
  on Knowledge Discovery \& Data Mining}. ACM, 2019, pp. 1946--1956.

\bibitem{10.1371/journal.pmed.1002730}
Jakob~Nikolas Kather, Johannes Krisam, et~al.,
\newblock ``Predicting survival from colorectal cancer histology slides using
  deep learning: A retrospective multicenter study,''
\newblock {\em PLOS Medicine}, vol. 16, no. 1, pp. 1--22, 01 2019.

\bibitem{wang2017chestxray}
Xiaosong Wang, Yifan Peng, et~al.,
\newblock ``Chestx-ray8: Hospital-scale chest x-ray database and benchmarks on
  weakly-supervised classification and localization of common thorax
  diseases,''
\newblock in {\em CVPR}, 2017, pp. 3462--3471.

\bibitem{DBLP:journals/corr/abs-1803-10417}
Philipp Tschandl, Cliff Rosendahl, and Harald Kittler,
\newblock ``The ham10000 dataset, a large collection of multi-source
  dermatoscopic images of common pigmented skin lesions,''
\newblock {\em Scientific data}, vol. 5, pp. 180161, 2018.

\bibitem{codella2019skin}
Noel Codella, Veronica Rotemberg, Philipp Tschandl, M~Emre Celebi, Stephen
  Dusza, David Gutman, Brian Helba, Aadi Kalloo, Konstantinos Liopyris, Michael
  Marchetti, et~al.,
\newblock ``Skin lesion analysis toward melanoma detection 2018: A challenge
  hosted by the international skin imaging collaboration (isic),''
\newblock {\em arXiv preprint arXiv:1902.03368}, 2019.

\bibitem{KERMANY20181122}
Daniel~S. Kermany, Michael Goldbaum, et~al.,
\newblock ``Identifying medical diagnoses and treatable diseases by image-based
  deep learning,''
\newblock {\em Cell}, vol. 172, no. 5, pp. 1122 -- 1131.e9, 2018.

\bibitem{deepdr}
DeepDR Diabetic Retinopathy Image~Dataset (DeepDRiD),
\newblock ``The 2nd diabetic retinopathy – grading and image quality
  estimation challenge,'' \url{https://isbi.deepdr.org/data.html}, 2020.

\bibitem{ALDHABYANI2020104863}
Walid Al-Dhabyani, Mohammed Gomaa, Hussien Khaled, and Aly Fahmy,
\newblock ``Dataset of breast ultrasound images,''
\newblock {\em Data in Brief}, vol. 28, pp. 104863, 2020.

\bibitem{DBLP:journals/corr/abs-1901-04056}
Patrick Bilic, Patrick~Ferdinand Christ, et~al.,
\newblock ``The liver tumor segmentation benchmark (lits),''
\newblock {\em CoRR}, vol. abs/1901.04056, 2019.

\bibitem{8625393}
X.~{Xu}, F.~{Zhou}, et~al.,
\newblock ``Efficient multiple organ localization in ct image using 3d region
  proposal network,''
\newblock {\em IEEE Transactions on Medical Imaging}, vol. 38, no. 8, pp.
  1885--1898, 2019.

\bibitem{pedregosa2011scikit}
Fabian Pedregosa, Ga{\"e}l Varoquaux, et~al.,
\newblock ``Scikit-learn: Machine learning in python,''
\newblock {\em the Journal of machine Learning research}, vol. 12, pp.
  2825--2830, 2011.

\bibitem{chollet2015keras}
Fran\c{c}ois Chollet et~al.,
\newblock ``Keras,'' \url{https://keras.io}, 2015.

\end{thebibliography}

\end{document}